\documentclass[sigconf,nonacm]{acmart}

\usepackage{tcolorbox}
\usepackage{graphicx} 
\usepackage{caption} 
\usepackage{booktabs}
\usepackage{subcaption}
\usepackage{listings}
\usepackage{svg}
\usepackage{multirow}

\setcopyright{acmlicensed}
\copyrightyear{2024}
\acmYear{2024}
\acmDOI{XXXXXXX.XXXXXXX}

\acmConference[UDM KDD '24]{Make sure to enter the correct
  conference title from your rights confirmation email}{August 25-26,
  2024}{Barcelona, Spain}
\acmISBN{978-1-4503-XXXX-X/18/06}

\begin{document}

\title{Measuring Aleatoric and Epistemic Uncertainty in LLMs: Empirical Evaluation on ID and OOD QA Tasks}


\author{Kevin Wang}
\authornote{Equal Contribution}
\affiliation{%
  \institution{The University of Texas at Dallas}
  \city{Richardson}
  \state{Texas}
  \country{USA}}
\email{kevin.wang@utdallas.edu}

\author{Subre Moktar}
\authornotemark[1]
\affiliation{%
  \institution{University of Texas at Dallas}
  \city{Richardson}
  \state{Texas}
  \country{USA}}
\email{subre.moktar@utdallas.edu}

\author{Jia Li}
\authornotemark[1]
\affiliation{%
 \institution{University of Texas at Dallas}
 \city{Richardson}
 \state{Texas}
 \country{USA}}
\email{jia.li@utdallas.edu}

\author{Kangshuo Li}
\authornotemark[1]
\affiliation{%
  \institution{University of Texas at Dallas}
  \city{Richardson}
  \state{Texas}
  \country{USA}}
\email{kangshuo.li@utdallas.edu}

\author{Feng Chen}
\affiliation{%
  \institution{University of Texas at Dallas}
  \city{Richardson}
  \state{Texas}
  \country{USA}}
\email{feng.chen@utdallas.edu}

\renewcommand{\shortauthors}{}

\begin{abstract}

Large Language Models (LLMs) have become increasingly pervasive, finding applications across many industries and disciplines. Ensuring the trustworthiness of LLM outputs is paramount, where uncertainty estimation (UE) plays a key role. 
In this work, a comprehensive empirical study is conducted to examine the robustness and effectiveness of diverse UE measures regarding aleatoric and epistemic uncertainty in LLMs. It involves twelve different UE methods and four generation quality metrics including LLMScore from LLM criticizers to evaluate the uncertainty of LLM-generated answers in Question-Answering (QA) tasks on both in-distribution (ID) and out-of-distribution (OOD) datasets.  
Our analysis reveals that information-based methods, which leverage token and sequence probabilities, perform exceptionally well in ID settings due to their alignment with the model's understanding of the data. Conversely, density-based methods and the P(True) metric exhibit superior performance in OOD contexts, highlighting their effectiveness in capturing the model's epistemic uncertainty. Semantic consistency methods, which assess variability in generated answers, show reliable performance across different datasets and generation metrics. These methods generally perform well but may not be optimal for every situation.
\end{abstract}



\maketitle

\section{Introduction}




In recent years, Large Language Models (LLMs) have exploded in popularity and performance. Nowadays, we see evidence of the usage of LLMs in a wide range of fields, from sciences to arts and in between. The increased utilization of these models prompts research questions: How can we assess the disentangled uncertainties in the responses they produce? Furthermore, how does the reliability of uncertainty estimates in these responses change when we encounter out-of-distribution questions?



There are various uncertainty estimation approaches to evaluating whether a model generates trustworthy understanding and answers if the model comprehends the question\cite{liu2023trustworthy, guo2023evaluating}. 
This can be done in several ways, such as asking the same prompt multiple times and analyzing the semantic consistency of the responses\cite{fisch2022uncertainty, fadeeva2023lm, kuhn2023semantic, raj2023semantic}. 
However, assessing the truthfulness of model responses is also crucial. Several studies have proposed metrics that assess the effectiveness of uncertainty methods, specifically integrating the evaluation of both uncertainty and correctness, where the average quality of answers is expected to be high when filtering out uncertain responses with the uncertainty estimates\cite{fadeeva2023lm, malinin2017incorporating}.
Notably, the correctness of model responses is measured using metrics such as Rouge-L or BERTScore under a supervised QA setting \cite{zhang2020bertscore}. With the continuous development of LLMs, we also consider LLM collaboration as a feasible approach to utilize extra LLM apart from the question-answer generator to give quality scores for evaluation \cite{feng2024don, wang2024evaluating}. 

To address the initial questions posed by this study, we collect extensive empirical results to explore the differences and pitfalls in terms of the effectiveness of four classes of uncertainty metrics with both ID and OOD datasets for supervised QA tasks over four quality measures. Detailed interpretations are made based on uncertainties' formulation.
Further, we also focus on two types of uncertainty that can cause these hallucinations: aleatoric and epistemic uncertainty. Aleatoric uncertainty represents the randomness inherent during generation, which we assess using in-distribution data where the model is expected to know the answer but may still output wrong answers due to randomness. Epistemic uncertainty represents a lack of knowledge, assessed using out-of-distribution data where the model has no prior exposure to the information and thus is likely to produce incorrect responses. Comparisons of the effectiveness ranking between ID and OOD datasets also imply the heterogeneous adaptability for these types of methods.

To summarize, this study provides a rigorous evaluation of UE methods across ID datasets and an OOD dataset based on generations from Llama-2-7b-chat \cite{touvron2023llama}, investigating their capacity to manage aleatoric and epistemic uncertainties. Additionally, we explored using different LLMs to generate correctness responses for the uncertainty quantification, specifically the Gemma-7b-it \cite{gemma_2024} model and the Vicuna-7b-v1.5 model \cite{vicuna2023}. Our findings indicate that while semantic consistency methods are generally reliable, they are not always the most effective. Information-based methods demonstrate robust performance in ID tasks, effectively handling aleatoric uncertainty. In contrast, density methods and P(True) measures offer insightful perspectives on OOD scenarios, more adeptly addressing epistemic uncertainty. Collectively, our analysis reveals that no single UE method universally excels; the selection of an appropriate method is contingent upon the specific uncertainty type and data context. This nuanced understanding allows us to delineate the capabilities of LLMs in handling diverse data challenges, offering a structured framework for applying UE methods in practical settings. A detailed discussion of these uncertainty estimation techniques is provided in Section \ref{sec:UEMethods} and their integration with accuracy assessments is provided in Section \ref{sec:Experiments}.




\vspace{-3mm}

\section{UE Methods}\label{sec:UEMethods}

\textbf{Semantic Consistency Responses:} These uncertainty estimation (UE) methods leverage the spreading of meaning of responses and classify them accordingly. The first method, Semantic Entropy \cite{kuhn2023semantic}, utilizes the DeBERTa model to generate a diverse set of responses. It then assesses the similarity between each response. When responses are consistently classified into the same category, it indicates a high degree of certainty from the model regarding its answers. The Eigenvalue Laplacian method tests the semantic sets using Natural Language Inference (NLI) scores to measure the semantic similarity among the responses \cite{lin2024generating}. Eccentricity involves using embeddings and calculating the norm between the eigenvectors of the responses to measure their semantic distance \cite{lin2024generating}. Lexical Similarity calculates the mean similarity between all pairs of generated samples to assess the similarity of the responses in terms of word choice and phrasing \cite{10.1162/tacl_a_00330}. Lastly, the Degree Matrix determines the pairwise distance between the sums of the similarities of the responses to assess their overall similarity \cite{lin2024generating}. Each of these methods provides a different perspective on the consistency and similarity of the generated responses. 

\textbf{Information-Based Responses:} These UE methods focus more on the generation of the text and calculate a conditional probability for each response. Perplexity employs the greedy log-likelihoods to generate a value \cite{10.1162/tacl_a_00330} and then finds the means of the probability in the sentence.
Mean Token Entropy estimates the mean entropy of every token in the generation \cite{10.1162/tacl_a_00330}. Maximum Sequence Probability uses the log probability of each token and then finds the sum \cite{fadeeva2023lm}. Monte Carlo Sequence Entropy applies Monte Carlo estimates to determine uncertainty, as explored in \cite{kuhn2023semantic}.

\textbf{Density-Based Responses:} These UE methods check the density of the responses. Mahalanobis Distance Sequence Decoder \cite{NEURIPS2018_abdeb6f5} calculates the covariance matrix, then determines the difference from the center. Robust Density Estimation \cite{yoo-etal-2022-detection} is similar to Mahalanobis distance but involves dimension reduction and calculating the covariance matrix in a more robust way.

\textbf{Reflexive Responses:} 
The method known as P(True) \cite{kadavath2022language} simply queries the model that generated the answer to verify the correctness of its responses. 


We notice several limitations introduced by the LM-Polygraph framework. One is the choice of quality evaluation metrics that are only related to text similarity or common subsequences, where we introduce LLM collaboration to utilize heterogeneous knowledge to get quality evaluations with 2 different backbones \cite{feng2024don}. LLM criticizers also show more aligned behavior compared with human evaluation. Additionally, our study novelly disentangles aleatoric uncertainty and epistemic uncertainty and dives into more specific application scenarios where the LLM might lack knowledge about questions with novel context, and an explanation of various UE effectiveness variations is discussed based on our empirical results.


\vspace{-1mm}

\section{Experiments}\label{sec:Experiments}

We adapted the framework of the LM-Polygraph \cite{fadeeva2023lm}, incorporating new datasets and introducing a novel generation metric referred to as LLMScore. This integration allows for the collection of generation metrics from LLMs, facilitating a comprehensive analysis. Drawing inspiration from previous collaborative efforts in LLM evaluation such as QA-Eval \cite{wang2024evaluating}, our approach provides an alternative method for conducting thorough evaluations. 

\vspace{-3mm}

\subsection{Empirical Settings}
\textbf{Datasets:}      The datasets employed in this paper include CoQA \cite{reddy2019coqa}, bAbIQA \cite{dodge2016evaluating}, and ALCUNA \cite{yin2023alcuna}. CoQA and bAbIQA are widely recognized datasets commonly utilized for evaluating model performance in various tasks and mainly focusing on the ability for a model to generate an answer. CoQA focuses on assessing a model's ability to comprehend stories while engaging in conversation, whereas bAbIQA evaluates the reading comprehension capabilities.

In contrast, ALCUNA presents an out-of-distribution challenge by generating artificial entities based on existing entity attributes, which is guaranteed to be out of the training corpus. Consequently, this dataset serves to test the model's adaptability to encountering new information. We only selected the free response questions so as to test our LLM on tasks similar to the prior two datasets.

\textbf{Model:}     This paper employed the Llama-2-7b-chat \cite{touvron2023llama} model from Hugging Face. We selected this model due to its established standardization during our study period, enabling us to directly compare our findings with those presented in the LM-Polygraph paper.

\textbf{Prompts:}   We use tailored prompts for each QA task to fit in different QA scenarios. For CoQA, the prompt for answer generation guides the LLM to utilize in-context learning characteristics:
\vspace{-2mm}

\begin{tcolorbox}[colback=white,colframe=darkgray,boxrule=0.1mm,left=1mm,right=1mm,top=1mm,bottom=1mm]
\scriptsize
\texttt{Answer the question given a story. Output only the answer itself. \\
Story: \textbf{[Story]} \\
Question: \textbf{[Question]} \\
Answer:
}
\end{tcolorbox}

\vspace{-2mm}
For bAbIQA, we used one-shot prompting strategy and added more instructional requirements, pruning redundant responses while formatting the answers that align with the golden answer: 
\vspace{-2mm}
\begin{tcolorbox}[colback=white,colframe=darkgray,boxrule=0.1mm,left=1mm,right=1mm,top=1mm,bottom=1mm]
\scriptsize
\texttt{Imagine that you are only able to say a single word. Answer the question given a context. You must only output the full name of the location the same way it is mentioned in the text. Do not try to be polite or helpful. \\
Example: \\
Context: Mary moved to the bathroom. John went to the hallway. Daniel went back to the hallway. Sandra moved to the garden. John moved to the office. Sandra journeyed to the bathroom. Mary moved to the hallway. Daniel traveled to the office. John went back to the garden. John moved to the bedroom. \\
Question: Where is Sandra? \\
Answer: bathroom \\
Context: \textbf{[Context]} \\
Question: \textbf{[Question]} \\
Answer: 
}
\end{tcolorbox}
\vspace{-2mm}

The generator LLM needs to be prompted differently when handling ALCUNA due to its OOD nature. We added one more option called "I don't know. " to handle OOD questions, where this option is assigned with the best quality of answer, implying the ability to be aware of high epistemic uncertainty:
\vspace{-2mm}
\begin{tcolorbox}
[colback=white,colframe=darkgray,boxrule=0.1mm,left=1mm,right=1mm,top=1mm,bottom=1mm]
\scriptsize
\texttt{Answer the question given. Output only the answer itself and if you don't know the answer only respond with 'I don't know'. ONLY RESPOND with the answer to the question. DO NOT repeat the question or include a greeting.  \\
Example: What is the latitude of the habitat of Cliteulata? \\
Answer: The Answer is: -78.5 degrees  \\
Question: \textbf{[Question]} \\
Answer:
}
\end{tcolorbox}
\vspace{-2mm}

\textbf{Generation Metrics:}        
To evaluate the correctness of answers generated by the LLM, we employed various generation metrics. These include Rouge-L, BERTScore \cite{zhang2020bertscore}, LLMScore, and manual human evaluations. LLMScore is our implementation of QA-Eval \cite{wang2024evaluating}, utilizing two different models: Vicuna-7b-v1.5 \cite{vicuna2023} and Google Gemma \cite{gemma_2024}. The scores are collected with the following template:

\vspace{-2mm}
\begin{tcolorbox}
[colback=white,colframe=darkgray,boxrule=0.1mm,left=1mm,right=1mm,top=1mm,bottom=1mm]
\scriptsize
\texttt{Here is a prompt, a golden answer, and an AI-generated answer. Can you judge how correct the AI-generated answer is according to the golden answer? Please return a quality score regarding semantic similarity. Reply only with a number between 0.000 and 1.000. DO NOT include anything other than the number in your response.\\
PROMPT: \textbf{[PROMPT]} \\
GOLDEN ANSWER: \textbf{[GOLDEN ANSWER]} \\
AI-GENERATED ANSWER: \textbf{[AI-GENERATED ANSWER]}
}
\end{tcolorbox}
\vspace{-2mm}

In our results, we primarily focus on the scores determined by LLMScore using Google Gemma. This choice is based on our observations that Gemma's evaluations closely align with manual human evaluations of a subsample from each dataset. Excluding bAbIQA, which we sampled 1000 examples from, each human evaluation was performed on a subsample of 600 examples. The human work was performed by students.

\par Our generation metrics operate by taking a single example as input, which includes the prompt, the LLM-generated answer, and the golden answer. The LLM-generated answer is denoted by $f(x_i)$. The generation evaluator outputs a score $Q(f(x_i), x_i, y_i)$ within a continuous range from 0 to 1, where 1 indicates a perfect semantic match between the LLM-generated answer and the golden answer, and 0 indicates no semantic similarity. 

\textbf{UE Method Scoring Metrics:}        The effectiveness of UE methods varies significantly. We assess a UE method based on its ability to determine whether an LLM-generated answer is acceptable or unacceptable. Ideally, the uncertainty score of an LLM-generated answer should be inversely correlated with its quality: high uncertainty should correlate with low quality and vice versa. This relationship can be quantified using the Prediction Rejection Ratio (PRR) \cite{malinin-etal-2017-incorporating}.

\par To calculate PRR, we first generate a prediction rejection (PR) curve by plotting the averaged quality of answers $Q(f(x_i), x_i, y_i)$  against its associated uncertainty estimate. The goal of this approach is to use the uncertainty of an answer to decide whether or not to "reject" it. Intuitively, rejection should occur when an answer is deemed to be highly uncertain, as this means that the answer is likely of low quality and unsuitable for the question. 

\par We then compute the PRR by finding the ratio between the area under the PR curve (AUCPR) calculated from the UE method in question and that from an oracle method: 
\[
\mathtt{PRR} = \frac{\mathtt{AUCPR}_{UE}}{\mathtt{AUCPR}_{ORACLE}}
\]

Here, the oracle curve represents optimal rejection while the UE curve represents rejection based on the uncertainty method in question. There exists a random curve which dictates a random order of rejection, as a baseline. Our PRR is then computed by finding the ratio between the area between the UE curve and random curve and the area between the oracle curve and random curve. A PRR close to 1 indicates that the UE method rejects close to the optimal amount of answers.
This ratio effectively indicates the utility of a UE method by measuring its usefulness in determining the correctness of an LLM's answer. A higher PRR signifies a more effective UE method, while a lower PRR indicates a less effective method for guaranteeing the quality of LLM-generated answers. 
For our experiments, we used Nvidia A100 80GB GPUs.


\subsection{Empirical Results}
Both ID and OOD datasets revealed that semantic consistency methods are generally reliable but not always optimal. Information-based methods are more suitable for ID tasks, whereas density methods and P(True) provide valuable insights into OOD uncertainty. Tables \ref{tab:CoQA-bAbIQA Results}, \ref{tab:ALCUNA Results} show the PRR results across the CoQA, bAbIQA, and ALCUNA datasets, respectively. CoQA and bAbIQA represent ID datasets, with ALCUNA being an OOD dataset. Our results therefore span the contexts of both aleatoric and epistemic uncertainty. This allows us to examine the differences in efficacies of UE metrics across the two types of uncertainty and make conclusions on the results.
\par
\begin{figure}
    \centering
    \includegraphics[width=0.9\linewidth]{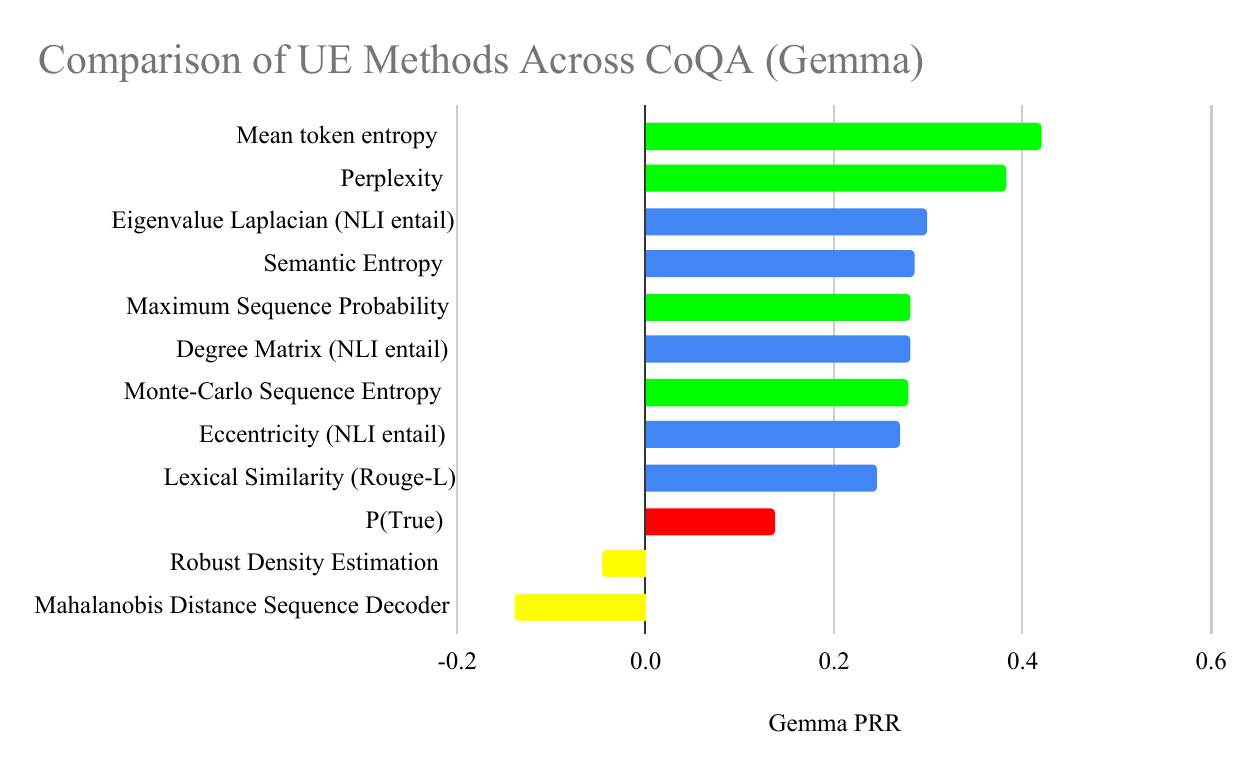}
    \vspace{-2mm}
    \caption{Ranking of UE Methods on CoQA, using LLMScore with Gemma.}
    \label{fig:UE Method Ranking CoQA}
    \vspace{-3mm}
\end{figure}

\begin{figure}
    \centering
    \includegraphics[width=0.9\linewidth]{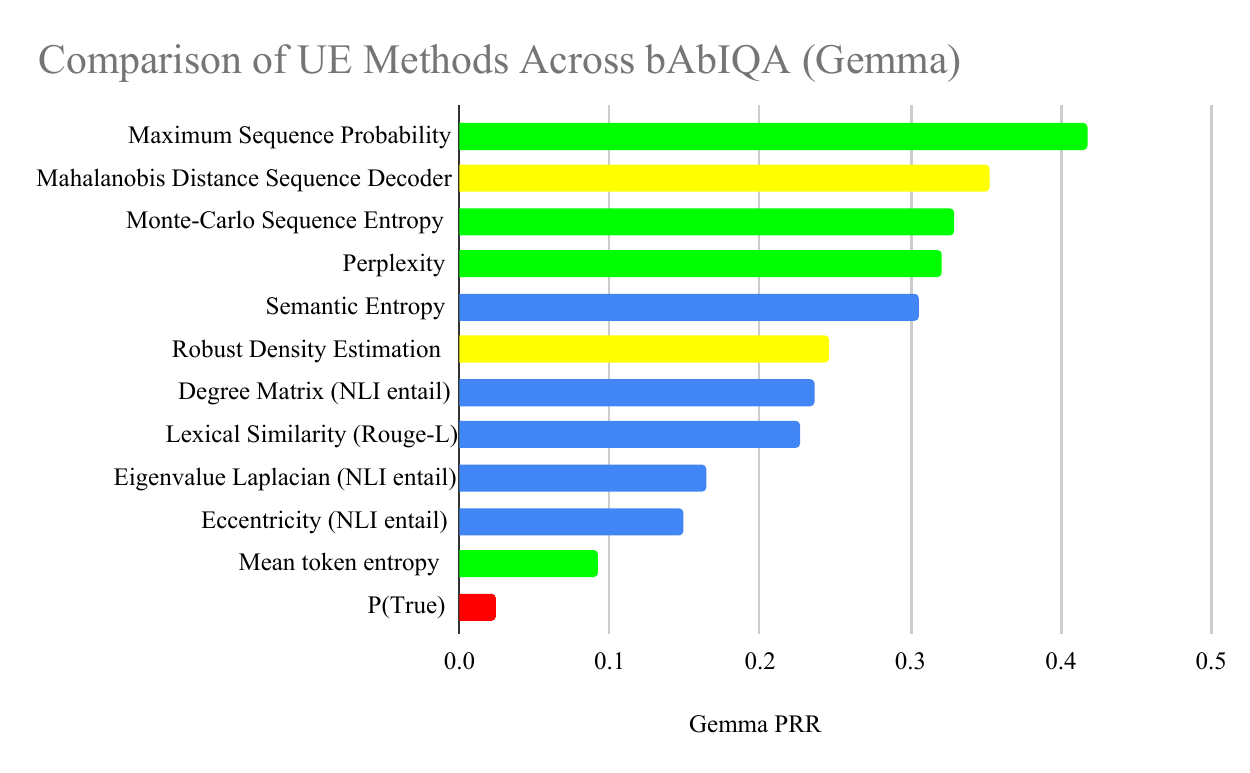}
    \vspace{-2mm}
    \caption{Ranking of UE Methods on bAbIQA, using LLMScore with Gemma.}
    \label{fig:UE Method Ranking bAbIQA}
    \vspace{-3mm}
\end{figure}

\begin{figure}
    \centering
    \includegraphics[width=0.9\linewidth]{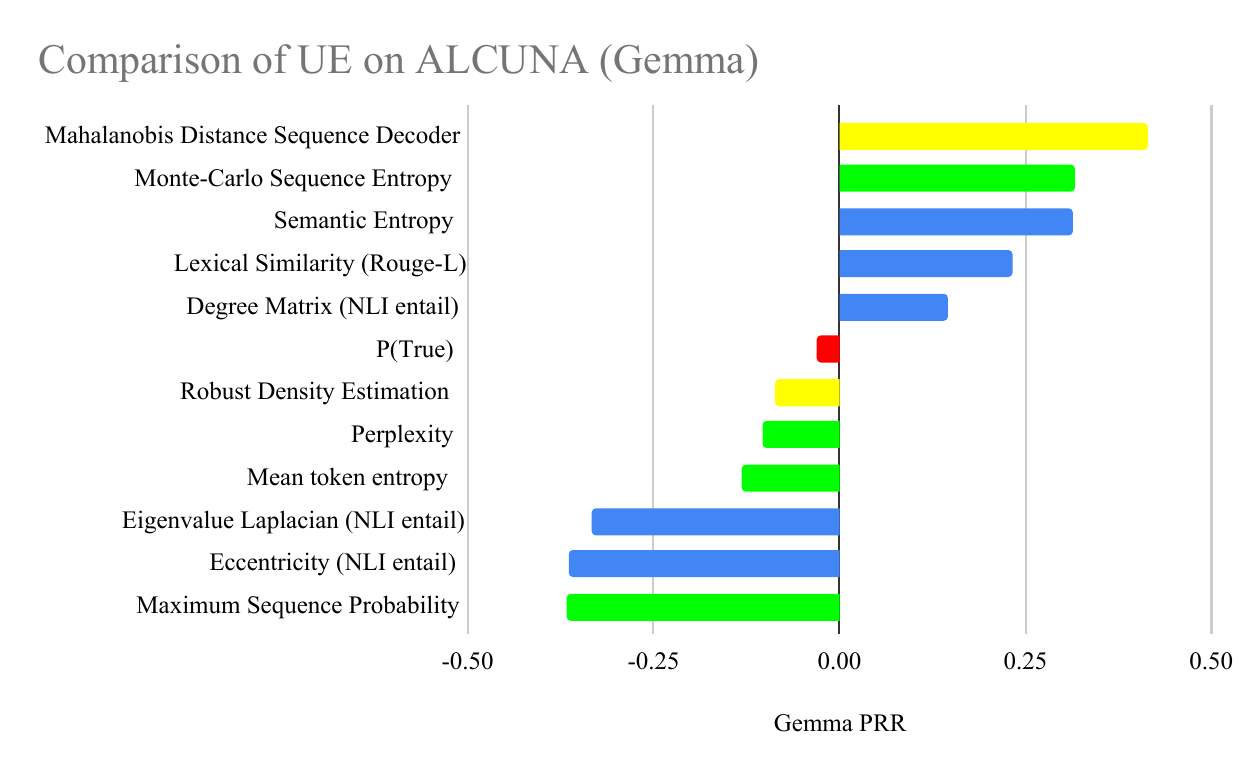}
    \vspace{-2mm}
    \caption{Ranking of UE Methods on ALCUNA, using LLMScore with Gemma.}
    \label{fig:UE Method Ranking ALCUNA}
    \vspace{-3mm}
\end{figure}

 Figures \ref{fig:UE Method Ranking CoQA} and \ref{fig:UE Method Ranking bAbIQA} show the rankings of UE methods by PRR for CoQA and bAbIQA, our ID datasets. Figure \ref{fig:UE Method Ranking ALCUNA} shows the UE rankings for ALCUNA, our OOD dataset. For these figures, we present our findings using LLMScore with Gemma as the model backbone. Different colors in the figures represent different classes of UE methods, with green as information-based methods, blue as semantic consistency methods, yellow as density-based methods, and red as reflexive methods.
\par It is immediately noticeable that these three rankings of UE methods have some similarities and some distinctions. In general, we observe that the semantic consistency UE methods (shown in blue) seem often serviceable but not necessarily optimal for both aleatoric and epistemic uncertainty. Subsequently, we will discuss our findings for aleatoric and epistemic uncertainty individually.

\begin{table*}[ht]
    \centering
    \footnotesize
    \caption{PRR scores on the CoQA and bAbIQA datasets using Llama-2-7b-chat with various UE methods and generation metrics. Bold represents best in column.}
    \vspace{-4mm}
    
    \resizebox{\textwidth}{!}{%
    \begin{tabular}{c|c|c|c|c|c|c|c|c|c|c} 
    \toprule
    & \multicolumn{5}{c|}{CoQA} & \multicolumn{5}{c}{bAbIQA} \\ \midrule
    
    \multirow{2}{*}{\textbf{UE Method}} & \multirow{2}{*}{\textbf{ROUGE-L}} & \multirow{2}{*}{\textbf{BERTScore}}  & \multicolumn{2}{c|}{\textbf{LLMScore}} & \multirow{2}{*}{\textbf{HumanEval}} & \multirow{2}{*}{\textbf{ROUGE-L}} & \multirow{2}{*}{\textbf{BERTScore}}  & \multicolumn{2}{c|}{\textbf{LLMScore}} & \multirow{2}{*}{\textbf{HumanEval}}  \\ \cline{4-5} \cline{9-10}
    
    & & & \textbf{Gemma} & \textbf{Vicuna} &  & & &\textbf{Gemma} & \textbf{Vicuna} &  \\ \hline 
    
    Maximum Sequence Probability & 0.398 & 0.409 & 0.28 & -0.089 & 0.155 & 0.074 & 0.884 & \textbf{0.418} & 0.065 & \textbf{0.575} \\ \hline
    Semantic Entropy & 0.399 & 0.398 & 0.285 & 0.12 & \textbf{0.18} & 0.53 & 0.86 & 0.305 & 0.214 & 0.534 \\ \hline
    Mahalanobis Distance Sequence Decoder & -0.044 & 0.048 & -0.139 & -0.237 & -0.012 & 0.437 & 0.121 & 0.352 & \textbf{0.373} & 0.317 \\ \hline
    Robust Density Estimation & \textbf{0.403} & 0.401 & -0.046 & -0.192 & -0.044 & 0.345 & 0.201 & 0.246 & 0.262 & 0.571 \\ \hline
    Perplexity & 0.255 & 0.095 & 0.381 & 0.166 & 0.159 & 0.575 & 0.95 & 0.32 & 0.217 & 0.468 \\ \hline
    Mean Token Entropy & 0.287 & 0.106 & \textbf{0.419} & 0.191 & 0.168 & 0.064 & 0.9 & 0.092 & 0.082 & 0.424 \\ \hline
    P(True) & -0.098 & -0.256 & 0.136 & \textbf{0.432} & -0.026 & 0.08 & 0.058 & 0.024 & -0.029 & -0.375 \\ \hline
    Eccentricity (NLI entail) & 0.399 & 0.393 & 0.268 & -0.05 & 0.122 & 0.148 & 0.619 & 0.148 & 0.187 & 0.511 \\ \hline
    Eigenvalue Laplacian (NLI entail) & 0.393 & 0.412 & 0.297 & -0.08 & 0.129 & 0.191 & 0.623 & 0.163 & 0.199 & 0.433 \\ \hline
    Lexical Similarity (Rouge-L) & 0.401 & \textbf{0.426} & 0.245 & -0.079 & 0.09 & 0.443 & 0.872 & 0.226 & 0.299 & 0.442 \\ \hline
    Monte-Carlo Sequence Entropy & 0.395 & 0.401 & 0.277 & -0.078 & 0.146 & \textbf{0.587} & \textbf{0.962} & 0.328 & 0.364 & 0.57 \\ \hline
    Degree Matrix (NLI entail) & 0.355 & 0.336 & 0.279 & -0.087 & 0.173 & 0.475 & 0.882 & 0.236 & 0.321 & 0.483 \\ 
    \bottomrule
    \end{tabular}
    }
    \label{tab:CoQA-bAbIQA Results}
\end{table*}

\begin{table*}[ht]
    \centering
    \footnotesize
    \caption{PRR scores on the ALCUNA dataset using Llama-2-7b-chat with various UE methods and generation metrics. Bold represents best in column.}
    \vspace{-4mm}
    \begin{tabular}{c|c|c|c|c|c} \toprule 
 \multicolumn{6}{c}{ALCUNA}\\ \midrule
 \textbf{UE Method}& \textbf{ROUGE-L}& \textbf{BERTScore}& \textbf{LLMScore (Gemma)}& \textbf{LLMScore (Vicuna)}& \textbf{HumanEval}\\ \hline 
         Maximum Sequence Probability
&  0.607&  \textbf{0.783}&  -0.368&  0.113&  0.512
\\ \hline  
         Semantic Entropy&  0.625&  0.602&  0.313&  -0.499&  0.647
\\ \hline  
         Mahalanobis Distance Sequence Decoder&  -0.242&  -0.183&  \textbf{0.414}&  \textbf{0.479}&  0.334
\\ \hline  
         Robust Density Estimation&  -0.123&  -0.162&  -0.085&  0.121&  0.395
\\ \hline  
         Perplexity
&  0.119&  0.022&  -0.103&  -0.647&  0.658
\\ \hline  
         Mean Token Entropy&  \textbf{0.728}&  0.628&  -0.131&  -0.163&  \textbf{0.711}\\ \hline  
         P(True)&  0.11&  0.012&  -0.032&  -0.752&  0.334
\\ \hline  
         Eccentricity (NLI entail)
&  0.047&  0.294&  -0.365&  -0.236&  0.643
\\ \hline  
         Eigenvalue Laplacian (NLI entail)&  0.072&  0.299&  -0.331&  -0.408&  0.645
\\ \hline  
         Lexical Similarity (Rouge-L)&  0.282&  0.353&  0.231&  0.058&  0.252
\\ \hline  
 Monte-Carlo Sequence Entropy& 0.543& 0.584& 0.315& -0.202& 0.542
\\ \hline  
 Degree Matrix (NLI entail)& 0.552& 0.413& 0.145& -0.586& 0.614\\ \bottomrule
    \end{tabular}
\vspace{-3mm}
\label{tab:ALCUNA Results}
\end{table*}

\subsubsection{Aleatoric Uncertainty Estimation}
Aleatoric uncertainty estimation is essential for understanding the variability inherent in the data used to train language models. For In-Distribution (ID) datasets, such as CoQA and bAbIQA, several UE methods were evaluated to measure their effectiveness in estimating aleatoric uncertainty.

From our results, we observed that information-based methods (green) tend to perform better on ID datasets. Specifically, Mean Token Entropy and Perplexity scored the highest on CoQA, with Gemma as the evaluation model. These methods likely excel due to their reliance on token probabilities, which align well with the structured nature of ID tasks. Similarly, methods like Semantic Entropy and Maximum Sequence Probability also showed strong performance on CoQA, indicating their robustness in handling token and sequence probabilities.

On the other hand, for bAbIQA, Maximum Sequence Probability emerged as the most effective method, followed by Mahalanobis Distance Sequence Decoder and Monte-Carlo Sequence Entropy. The nature of bAbIQA’s single-word answers might contribute to the effectiveness of these density-based methods (yellow). Interestingly, Mean Token Entropy, which performed well on CoQA, showed poor performance on bAbIQA, highlighting the variability in method effectiveness across different ID datasets.

\begin{figure}
        \centering
            \includegraphics[width=0.9\linewidth]{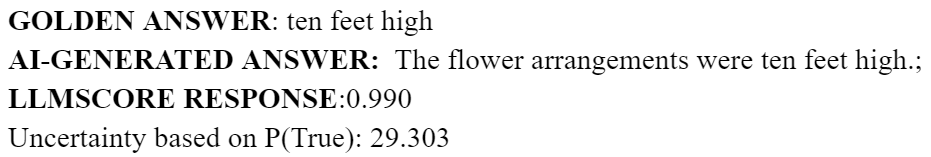}
        \vspace{-2mm}
        \caption{An example of P(True) estimating high uncertainty on a correct LLM answer in CoQA. 29.303 is considered a high value among the values collected.}
        \label{fig:CoQA Case Study}
        \vspace{-5mm}
\end{figure}

P(True) (red) consistently performed poorly across both CoQA and bAbIQA. This suggests that models struggle to accurately assess their own aleatoric uncertainty (see Fig. \ref{fig:CoQA Case Study}). Additionally, the performance of Vicuna and Gemma models varied: Gemma generally outperformed Vicuna on CoQA, while Vicuna showed better results for some density-based estimators on bAbIQA. This discrepancy could be attributed to Vicuna being a fine-tuned version of the base model Llama-2. Related work has found that RLHF-treated models tend to exhibit biases and reduced diversity, which could affect the performance of Vicuna as a generation evaluator \cite{kirk2024understanding}. 


\vspace{-1mm}

Overall, our findings indicate that information-based methods are more effective for ID datasets, while density-based methods perform variably depending on the dataset characteristics. Rouge-L and BERTScore also tend to score higher than LLMScore and HumanEval in PRR, suggesting that they are more closely aligned with uncertainty estimation for ID tasks.

\subsubsection{Epistemic Uncertainty Estimation}
Epistemic uncertainty estimation deals with the uncertainty arising from the model's lack of knowledge about the data, making it crucial for out-of-distribution (OOD) datasets like ALCUNA. Our evaluation of UE methods on ALCUNA revealed distinct patterns compared to ID datasets.

Among the UE methods, Mahalanobis Distance Sequence Decoder and Monte-Carlo Sequence Entropy demonstrated the highest effectiveness, similar to their performance on bAbIQA. These methods are likely more robust in OOD contexts due to their ability to generalize beyond the training data. Semantic Entropy also consistently ranked in the top four methods for ALCUNA, indicating its reliability across different datasets and both ID and OOD contexts.

\begin{figure}
        \centering
        \includegraphics[width=0.9\linewidth]{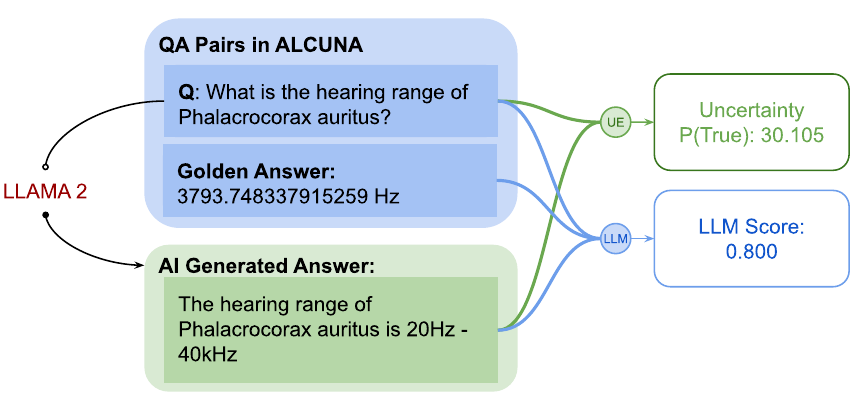}
        \vspace{-2mm}
        \caption{An example of P(True) estimating high uncertainty on an incorrect LLM answer in ALCUNA despite LLMScore indicating high correctness. }
        \label{fig:ALCUNA Case Study}
        \vspace{-5mm}
\end{figure}

P(True) (red) performed significantly better on ALCUNA than on the ID datasets (see Fig. \ref{fig:ALCUNA Case Study}). While it still cannot be considered an effective method, it outperformed many more methods than it did in ID contexts. This suggests that LLMs may be better at estimating their own epistemic uncertainty than aleatoric uncertainty.


Interestingly, the effectiveness of generation metrics varied significantly. BERTScore and Rouge-L, while generally high-scoring for ID datasets, showed a mixed performance on OOD datasets. Semantic Entropy's consistent performance across CoQA and ALCUNA, despite being less effective on bAbIQA, underscores the importance of selecting appropriate UE methods based on dataset characteristics.

The variability in method performance between ID and OOD datasets underscores the need for tailored approaches in uncertainty quantification. While density-based methods excel in representing training data and generalizing to OOD contexts, their effectiveness can vary based on the generation metrics used. For instance, these methods performed better when paired with Gemma and Vicuna, but worse with BERTScore and Rouge-L.

\vspace{-1mm}

\section{Conclusion}

As evident, there is no definitive best uncertainty estimation (UE) method. The suitability of different methods varies depending on the nature of the data. Information-based methods appear to be most effective for addressing aleatoric uncertainty, while density-based methods are better suited for epistemic uncertainty.

\vspace{-1mm}

\section{Future Work}
In future work, we intend to further investigate these distinctions using a more recent model, such as llama-3-8b-Instruct, instead of llama-2-7b-chat. Additionally, we aim to identify and incorporate more out-of-distribution (OOD) datasets for comparison, as our current study was limited to a single OOD dataset. We also plan to explore the impact of pre-training a model specifically for tasks like question answering and examine whether there is a significant difference in the scores collected between pre-trained and non-pre-trained models.









\bibliographystyle{ACM-Reference-Format}
\bibliography{ref}


\begin{thebibliography}{23}


\ifx \showCODEN    \undefined \def \showCODEN     #1{\unskip}     \fi
\ifx \showISBNx    \undefined \def \showISBNx     #1{\unskip}     \fi
\ifx \showISBNxiii \undefined \def \showISBNxiii  #1{\unskip}     \fi
\ifx \showISSN     \undefined \def \showISSN      #1{\unskip}     \fi
\ifx \showLCCN     \undefined \def \showLCCN      #1{\unskip}     \fi
\ifx \shownote     \undefined \def \shownote      #1{#1}          \fi
\ifx \showarticletitle \undefined \def \showarticletitle #1{#1}   \fi
\ifx \showURL      \undefined \def \showURL       {\relax}        \fi
\providecommand\bibfield[2]{#2}
\providecommand\bibinfo[2]{#2}
\providecommand\natexlab[1]{#1}
\providecommand\showeprint[2][]{arXiv:#2}

\bibitem[et~al.(2023a)]%
        {vicuna2023}
\bibfield{author}{\bibinfo{person}{Chiang et al.}} \bibinfo{year}{2023}\natexlab{a}.
\newblock \bibinfo{title}{Vicuna: An Open-Source Chatbot Impressing GPT-4 with 90\%* ChatGPT Quality}.
\newblock
\urldef\tempurl%
\url{https://lmsys.org/blog/2023-03-30-vicuna/}
\showURL{%
\tempurl}


\bibitem[et~al.(2016)]%
        {dodge2016evaluating}
\bibfield{author}{\bibinfo{person}{Dodge et al.}} \bibinfo{year}{2016}\natexlab{}.
\newblock \bibinfo{title}{Evaluating Prerequisite Qualities for Learning End-to-End Dialog Systems}.
\newblock
\showeprint[arxiv]{1511.06931}~[cs.CL]


\bibitem[et~al.(2020)]%
        {10.1162/tacl_a_00330}
\bibfield{author}{\bibinfo{person}{Fomicheva et al.}} \bibinfo{year}{2020}\natexlab{}.
\newblock \showarticletitle{{Unsupervised Quality Estimation for Neural Machine Translation}}.
\newblock \bibinfo{journal}{\emph{Transactions of the Association for Computational Linguistics}}  \bibinfo{volume}{8} (\bibinfo{date}{09} \bibinfo{year}{2020}), \bibinfo{pages}{539--555}.
\newblock
\showISSN{2307-387X}
\href{https://doi.org/10.1162/tacl_a_00330}{doi:\nolinkurl{10.1162/tacl_a_00330}}
\showeprint{https://direct.mit.edu/tacl/article-pdf/doi/10.1162/tacl\_a\_00330/1923296/tacl\_a\_00330.pdf}


\bibitem[et~al.(2023b)]%
        {fadeeva2023lm}
\bibfield{author}{\bibinfo{person}{Fadeeva et al.}} \bibinfo{year}{2023}\natexlab{b}.
\newblock \showarticletitle{LM-polygraph: Uncertainty estimation for language models}.
\newblock \bibinfo{journal}{\emph{arXiv preprint arXiv:2311.07383}} (\bibinfo{year}{2023}).
\newblock


\bibitem[et~al.(2024a)]%
        {feng2024don}
\bibfield{author}{\bibinfo{person}{Feng et al.}} \bibinfo{year}{2024}\natexlab{a}.
\newblock \showarticletitle{Don't Hallucinate, Abstain: Identifying LLM Knowledge Gaps via Multi-LLM Collaboration}.
\newblock \bibinfo{journal}{\emph{arXiv preprint arXiv:2402.00367}} (\bibinfo{year}{2024}).
\newblock


\bibitem[et~al(2023)]%
        {guo2023evaluating}
\bibfield{author}{\bibinfo{person}{Guo et al}.} \bibinfo{year}{2023}\natexlab{}.
\newblock \showarticletitle{Evaluating large language models: A comprehensive survey}.
\newblock \bibinfo{journal}{\emph{arXiv preprint arXiv:2310.19736}} (\bibinfo{year}{2023}).
\newblock


\bibitem[et~al.(2022)]%
        {kadavath2022language}
\bibfield{author}{\bibinfo{person}{Kadavath et al.}} \bibinfo{year}{2022}\natexlab{}.
\newblock \showarticletitle{Language models (mostly) know what they know}.
\newblock \bibinfo{journal}{\emph{arXiv preprint arXiv:2207.05221}} (\bibinfo{year}{2022}).
\newblock


\bibitem[et~al.(2023a)]%
        {kuhn2023semantic}
\bibfield{author}{\bibinfo{person}{Kuhn et al.}} \bibinfo{year}{2023}\natexlab{a}.
\newblock \bibinfo{title}{Semantic Uncertainty: Linguistic Invariances for Uncertainty Estimation in Natural Language Generation}.
\newblock
\showeprint[arxiv]{2302.09664}~[cs.CL]


\bibitem[et~al.(2024b)]%
        {kirk2024understanding}
\bibfield{author}{\bibinfo{person}{Kirk et al.}} \bibinfo{year}{2024}\natexlab{b}.
\newblock \bibinfo{title}{Understanding the Effects of RLHF on LLM Generalisation and Diversity}.
\newblock
\showeprint[arxiv]{2310.06452}~[cs.LG]


\bibitem[et~al.(2023b)]%
        {liu2023trustworthy}
\bibfield{author}{\bibinfo{person}{Liu et al.}} \bibinfo{year}{2023}\natexlab{b}.
\newblock \showarticletitle{Trustworthy LLMs: a Survey and Guideline for Evaluating Large Language Models' Alignment}.
\newblock \bibinfo{journal}{\emph{arXiv preprint arXiv:2308.05374}} (\bibinfo{year}{2023}).
\newblock


\bibitem[et~al(2017a)]%
        {malinin2017incorporating}
\bibfield{author}{\bibinfo{person}{Malinin et al}.} \bibinfo{year}{2017}\natexlab{a}.
\newblock \showarticletitle{Incorporating uncertainty into deep learning for spoken language assessment}. In \bibinfo{booktitle}{\emph{Proceedings of the 55th Annual Meeting of the Association for Computational Linguistics (Volume 2: Short Papers)}}. \bibinfo{pages}{45--50}.
\newblock


\bibitem[et~al(2017b)]%
        {malinin-etal-2017-incorporating}
\bibfield{author}{\bibinfo{person}{Malinin et al}.} \bibinfo{year}{2017}\natexlab{b}.
\newblock \showarticletitle{Incorporating Uncertainty into Deep Learning for Spoken Language Assessment}. In \bibinfo{booktitle}{\emph{Proceedings of the 55th Annual Meeting of the Association for Computational Linguistics (Volume 2: Short Papers)}}. \bibinfo{publisher}{Association for Computational Linguistics}, \bibinfo{address}{Vancouver, Canada}, \bibinfo{pages}{45--50}.
\newblock
\href{https://doi.org/10.18653/v1/P17-2008}{doi:\nolinkurl{10.18653/v1/P17-2008}}


\bibitem[et~al.(2019)]%
        {reddy2019coqa}
\bibfield{author}{\bibinfo{person}{Reddy et al.}} \bibinfo{year}{2019}\natexlab{}.
\newblock \bibinfo{title}{CoQA: A Conversational Question Answering Challenge}.
\newblock
\showeprint[arxiv]{1808.07042}~[cs.CL]


\bibitem[et~al.(2023c)]%
        {touvron2023llama}
\bibfield{author}{\bibinfo{person}{Touvron et al.}} \bibinfo{year}{2023}\natexlab{c}.
\newblock \showarticletitle{Llama 2: Open foundation and fine-tuned chat models}.
\newblock \bibinfo{journal}{\emph{arXiv preprint arXiv:2307.09288}} (\bibinfo{year}{2023}).
\newblock


\bibitem[et~al.(2024c)]%
        {wang2024evaluating}
\bibfield{author}{\bibinfo{person}{Wang et al.}} \bibinfo{year}{2024}\natexlab{c}.
\newblock \showarticletitle{Evaluating open-qa evaluation}.
\newblock \bibinfo{journal}{\emph{Advances in Neural Information Processing Systems}}  \bibinfo{volume}{36} (\bibinfo{year}{2024}).
\newblock


\bibitem[Fisch et~al\mbox{.}(2022)]%
        {fisch2022uncertainty}
\bibfield{author}{\bibinfo{person}{Adam Fisch}, \bibinfo{person}{Robin Jia}, {and} \bibinfo{person}{Tal Schuster}.} \bibinfo{year}{2022}\natexlab{}.
\newblock \showarticletitle{Uncertainty estimation for natural language processing}. COLING.
\newblock


\bibitem[Gemma~Team(2024)]%
        {gemma_2024}
\bibfield{author}{\bibinfo{person}{Thomas Mesnard et~al. Gemma~Team}.} \bibinfo{year}{2024}\natexlab{}.
\newblock \showarticletitle{Gemma}.
\newblock  (\bibinfo{year}{2024}).
\newblock
\href{https://doi.org/10.34740/KAGGLE/M/3301}{doi:\nolinkurl{10.34740/KAGGLE/M/3301}}


\bibitem[Lee et~al\mbox{.}(2018)]%
        {NEURIPS2018_abdeb6f5}
\bibfield{author}{\bibinfo{person}{Kimin Lee}, \bibinfo{person}{Kibok Lee}, \bibinfo{person}{Honglak Lee}, {and} \bibinfo{person}{Jinwoo Shin}.} \bibinfo{year}{2018}\natexlab{}.
\newblock \showarticletitle{A Simple Unified Framework for Detecting Out-of-Distribution Samples and Adversarial Attacks}. In \bibinfo{booktitle}{\emph{Advances in Neural Information Processing Systems}}, Vol.~\bibinfo{volume}{31}. \bibinfo{publisher}{Curran Associates, Inc.}
\newblock
\urldef\tempurl%
\url{https://proceedings.neurips.cc/paper_files/paper/2018/file/abdeb6f575ac5c6676b747bca8d09cc2-Paper.pdf}
\showURL{%
\tempurl}


\bibitem[Lin et~al\mbox{.}(2024)]%
        {lin2024generating}
\bibfield{author}{\bibinfo{person}{Zhen Lin}, \bibinfo{person}{Shubhendu Trivedi}, {and} \bibinfo{person}{Jimeng Sun}.} \bibinfo{year}{2024}\natexlab{}.
\newblock \bibinfo{title}{Generating with Confidence: Uncertainty Quantification for Black-box Large Language Models}.
\newblock
\showeprint[arxiv]{2305.19187}~[cs.CL]


\bibitem[Raj et~al\mbox{.}(2023)]%
        {raj2023semantic}
\bibfield{author}{\bibinfo{person}{Harsh Raj}, \bibinfo{person}{Vipul Gupta}, \bibinfo{person}{Domenic Rosati}, {and} \bibinfo{person}{Subhabrata Majumdar}.} \bibinfo{year}{2023}\natexlab{}.
\newblock \showarticletitle{Semantic consistency for assuring reliability of large language models}.
\newblock \bibinfo{journal}{\emph{arXiv preprint arXiv:2308.09138}} (\bibinfo{year}{2023}).
\newblock


\bibitem[Yin et~al\mbox{.}(2023)]%
        {yin2023alcuna}
\bibfield{author}{\bibinfo{person}{Xunjian Yin}, \bibinfo{person}{Baizhou Huang}, {and} \bibinfo{person}{Xiaojun Wan}.} \bibinfo{year}{2023}\natexlab{}.
\newblock \bibinfo{title}{ALCUNA: Large Language Models Meet New Knowledge}.
\newblock
\showeprint[arxiv]{2310.14820}~[cs.CL]


\bibitem[Yoo et~al\mbox{.}(2022)]%
        {yoo-etal-2022-detection}
\bibfield{author}{\bibinfo{person}{KiYoon Yoo}, \bibinfo{person}{Jangho Kim}, \bibinfo{person}{Jiho Jang}, {and} \bibinfo{person}{Nojun Kwak}.} \bibinfo{year}{2022}\natexlab{}.
\newblock \showarticletitle{Detection of Adversarial Examples in Text Classification: Benchmark and Baseline via Robust Density Estimation}. In \bibinfo{booktitle}{\emph{Findings of the Association for Computational Linguistics: ACL 2022}}. \bibinfo{publisher}{Association for Computational Linguistics}, \bibinfo{address}{Dublin, Ireland}, \bibinfo{pages}{3656--3672}.
\newblock
\href{https://doi.org/10.18653/v1/2022.findings-acl.289}{doi:\nolinkurl{10.18653/v1/2022.findings-acl.289}}


\bibitem[Zhang et~al\mbox{.}(2020)]%
        {zhang2020bertscore}
\bibfield{author}{\bibinfo{person}{Tianyi Zhang}, \bibinfo{person}{Varsha Kishore}, \bibinfo{person}{Felix Wu}, \bibinfo{person}{Kilian~Q. Weinberger}, {and} \bibinfo{person}{Yoav Artzi}.} \bibinfo{year}{2020}\natexlab{}.
\newblock \bibinfo{title}{BERTScore: Evaluating Text Generation with BERT}.
\newblock
\showeprint[arxiv]{1904.09675}~[cs.CL]







\end{thebibliography}


\end{document}